\documentclass[runningheads,a4paper]{llncs}

\usepackage{amssymb}
\usepackage{graphicx}
\usepackage{diagbox}
\usepackage{booktabs}
\usepackage{amsmath}
\usepackage{float}
\usepackage{bm}
\usepackage{url}
\urldef{\mailsa}\path|{chenbd,nnzheng}@mail.xjtu.edu.cn|
\urldef{\mailsb}\path|{liangjunli}@xaut.edu.cn|
\urldef{\mailsc}\path|{principe}@cnel.ufl.edu|
\newcommand{\keywords}[1]{\par\addvspace\baselineskip
\newcommand{\bibFilename}{typeinst_cbd}
\noindent\keywordname\enspace\ignorespaces#1}

\begin{document}

\mainmatter  

\title{Kernel Least Mean Square with Adaptive Kernel Size}

%
%
\author{Badong Chen$^{1}$%
\thanks{Manuscript received Dec. 7, 2013. This work was supported by National NSF of China (No. 61372152). }%
\and Junli Liang$^{2}$ \and Nanning Zheng$^{1}$ \and Jos\'{e} C. Pr\'{i}ncipe$^{3}$}
\institute{$^1$The Institute of Artificial Intelligence and Robotics, \\
Xi¡¯an Jiaotong University, 28 Xianning West Road, Xi¡¯an 710049, China\\
\mailsa\\
$^2$ School of Automation \& Information Engineering, Xi'an University of Technology, Xi'an, China \quad \mailsb\\
$^3$ Department of Electrical and Computer Engineering, University of Florida, Gainesville, FL 32611 USA  \quad \mailsc}
%
%
\maketitle

\begin{abstract}
Kernel adaptive filters (KAF) are a class of powerful nonlinear filters developed in Reproducing Kernel Hilbert Space (RKHS). The Gaussian kernel is usually the default kernel in KAF algorithms, but selecting the proper kernel size (bandwidth) is still an open  important issue especially for learning with small sample sizes. In previous research, the kernel size was set manually or estimated in advance by Silverman¡¯s rule based on the sample distribution. This study aims to develop an online technique for optimizing the kernel size of the kernel least mean square (KLMS) algorithm. A sequential optimization strategy is proposed, and a new algorithm is developed, in which the filter weights and the kernel size are both sequentially updated by stochastic gradient algorithms that minimize the mean square error (MSE). Theoretical results on convergence are also presented. The excellent performance of the new algorithm is confirmed by simulations on static function estimation and short term chaotic time series prediction.
\end{abstract}

\section{Introduction}

Kernel based methods are successfully used in machine learning and nonlinear signal processing due to their inherent advantages of convex optimization and universality in the space of $L_2$ functions. By mapping the input data into a feature space associated with a Mercer kernel, many efficient nonlinear algorithms can be developed, thanks to the kernel trick.  Popular kernel methods include support vector machine (SVM) \cite{vapnik2000nature,scholkopflearning}, kernel regularization network \cite{girosi1995regularization}, kernel principal component analysis (KPCA) \cite{scholkopf1998nonlinear}, and kernel Fisher discriminant analysis (KFDA) \cite{yang2002kernel}, etc. These nonlinear algorithms show significant performance improvement over their linear counterparts.

Online kernel learning \cite{kivinen2004online,slavakis2008online,orabona2009bounded,zhao2011double} has also been extensively studied in the machine learning and statistical signal processing literature, and it provides efficient alternatives to approximate the desired nonlinearity incrementally. As the training data are sequentially presented to the learning system, online learning requires, in general, much less memory and computational cost. Recently, kernel based online algorithms for adaptive filtering have been developed and have become an emerging area of research \cite{principe2011kernel}. Kernel adaptive filters (KAF) are derived in Reproducing Kernel Hilbert Spaces (RKHS) \cite{aronszajn1950theory,burges1998tutorial}, by using the linear structure and inner product of this space to implement the well-established linear adaptive filtering algorithms that correspond to nonlinear filters in the original input space. Typical KAF algorithms include the kernel least mean square (KLMS) \cite{liu2008kernel,bouboulis2011extension}, kernel affine projection algorithms (KAPA) \cite{lius2008kernel}, kernel recursive least squares (KRLS) \cite{engel2004kernel}, and extended kernel recursive least squares (EX-KRLS) \cite{liu2009extended}, etc. With a radially symmetric Gaussian kernel they create a growing radial-basis function (RBF) network to learn the network topology and adapt free parameters directly from the training data. Among these KAF algorithms, the KLMS is the simplest, and fastest to implement yet very effective.

There are two main open challenges in the KAF algorithms . The first is their growing structure with each sample, which results in increasing computational costs and memory requirements especially in continuous adaptation scenarios. In order to curb the network growth  and to obtain a compact representation, a variety of sparsification techniques have been applied, where only the important input data are accepted as new centers. The presently available sparsification criteria include the novelty criterion \cite{platt1991resource}, approximate linear dependency (ALD) criterion \cite{engel2004kernel} surprise criterion \cite{liu2009information}, and so on. In a recent work \cite{chen2012quantized}, we have proposed a novel  method, the quantized kernel least mean square (QKLMS) algorithm, to compress the input space and hence constrain the network size  which is shown to be very effective in yielding a compact network with desirable accuracy.

Selecting a proper Mercer kernel is the second remaining problem that should be addressed when implementing kernel adaptive filtering algorithms, especially when the training data size is small. In this case, the kernel selection includes two parts: first, the kernel type is chosen, and second, its parameters are determined. Among various kernels, the Gaussian kernel is very popular and is usually a default choice in kernel adaptive filtering due to its universal approximating capability, desirable smoothness and numeric stability. The normalized Gaussian kernel is
\begin{equation}
\kappa \left( {u,u'} \right) = \exp \left( { - {{{{\left\| {\mathbf{u} - \mathbf{u}'} \right\|}^2}} \mathord{\left/
 {\vphantom {{{{\left\| {\mathbf{u} - \mathbf{u}'} \right\|}^2}} {2{\sigma ^2}}}} \right.
 \kern-\nulldelimiterspace} {2{\sigma ^2}}}} \right)
\end{equation}
where the free parameter $\sigma$($\sigma>0$) is called the kernel size (also known as the kernel bandwidth or smoothing parameter). In fact, the Gaussian kernel is strictly positive definite and as such produces a RKHS that is dense \cite{burges1998tutorial} and as such linear algorithms in this RKHS are universal approximators of smooth $L_2$ functions.  In principle this means that in the large sample size regime the asymptotic properties of the mean square approximation are independent of the kernel size $\sigma$ \cite{chen2012mean}. This means that the kernel size in KAF only affects the dynamics of learning, because in the initial steps  the sample size is always small, therefore both the accuracy for batch learning and the convergence properties for online learning are dependent upon the kernel size. This should be contrasted with the effect of the kernel size in classification where the kernel size controls both the accuracy and the generalization of the optimal solution \cite{vapnik2000nature,scholkopflearning}. Up to now, there are many methods for selecting a kernel size for the Gaussian kernel borrowed from the areas of statistics, nonparametric regression and kernel density estimation. The most popular methods for the selection of the kernel size are: cross-validation (CV) \cite{wahba1990spline,hardle1990applied,racine1993efficient,cawley2003efficient,an2007fast} which can always be used since the kernel size is a free parameter, penalizing functions \cite{hardle1990applied}, plug-in methods \cite{hardle1990applied,herrmann1997local}, Silverman¡¯s rule \cite{silverman1986density} and other rules of thumb \cite{jones1996brief}. The cross-validation, penalizing functions, and plug-in methods are computationally intensive and are not suitable for online kernel learning. The Silverman¡¯s rule is widely accepted in kernel density estimation although it is derived under a Gaussian assumption and is usually not appropriate for multimodal distributions. Besides the fixed kernel size, some adaptive or varying kernel size algorithms can also be found in the literature \cite{brunsdon1995estimating,katkovnik2002kernel,yuan2009adaptive,singh2011information}. This topic is also closely related to the techniques of \textit{multi-kernel learning} or \textit{learning the kernel} in the machine learning literature \cite{gonen2011multiple,herbster2004relative,ong2005learning,argyriou2005learning,jin2010online,orabona2012multi}. There the goal is typically to learn a combination of kernels based on some optimization methods, but in KAF this approach is normally avoided due to the computational complexity \cite{principe2011kernel}.

All the above mentioned methods, however, are not suitable for determining an optimal kernel size in online kernel adaptive filtering, since they either are batch mode methods or originate from a different problem, such as the kernel density estimation. Given that in online learning the number of samples is large and not specified a priori, the final solution will be practically independent of the kernel size. The real issue is therefore how to speed up convergence to the neighborhood of the optimal solution, which will also provide smaller network sizes. In the present work, by treating the kernel size as an extra parameter for the optimization, a novel sequential optimization framework is proposed for the KLMS algorithm. The new optimization paradigm allows for an online adaptation algorithm. At each iteration cycle, the filter weights and the kernel size are both sequentially updated to minimize the mean square error (MSE). As the kernel size is updated sequentially, the proposed algorithm is computationally very simple. The new algorithm can also be incorporated in the quantization method so as to yield a compact model.

The rest of the paper is organized as follows. In section II, we briefly revisit the KLMS algorithm. In section III, we propose a sequential optimization strategy for the kernel size in KLMS, and then derive a simple stochastic gradient algorithm to adapt the kernel size. In section IV, we give theoretical results on convergence of KLMS with adaptive kernel size. Specifically, we derive the energy conservation relation in RKHS, and on this basis we derive a sufficient condition for the mean square convergence, and arrive at the theoretical value of the steady-state excess mean-square error (EMSE). In section V, we present simulation examples on static function estimation and short term chaotic time series prediction to confirm the satisfactory performance of the KLMS with adaptive kernel size. Finally, in section VI, we present the conclusion.

\section{KLMS}

Suppose the goal is to learn a continuous input-output mapping $f:\mathbb{U}\to\mathbb{Y}$ based on a sequence of input-output examples (training data)$\left\{ {\mathbf{u}(i),y(i)} \right\}_{i = 1}^N$, where $\mathbb{U}\subset\mathbb{R}^m$ is the input domain, $\mathbb{Y}\subset\mathbb{R}$ is the desired output space. The hypothesis space for learning is assumed to be a Reproducing Kernel Hilbert Space (RKHS) $\mathcal{H}_k$ associated with a Mercer kernel $\kappa \left( {\mathbf{u},\mathbf{u}'} \right)$, a continuous, symmetric, and positive-definite function $\kappa: \mathbb{U} \times \mathbb{U} \to \mathbb{R}$ \cite{aronszajn1950theory}. To find such a function , one may solve the regularized least squares regression in $\mathcal{H}_k$:
\begin{equation}
\mathop {\min }\limits_{f \in {\mathcal{H}_k}} \sum\limits_{i = 1}^N {{{\left( {y(i) - f\left( {\mathbf{u}(i)} \right)} \right)}^2}}  + \gamma \left\| f \right\|_{{\mathcal{H}_k}}^2
\end{equation}
where ${\left\| . \right\|_{{\mathcal{H}_k}}}$ denotes the norm in $\mathcal{H}_k$, $\gamma  \ge 0$ is the regularization factor that controls the smoothness of the solution. As the inner product in RKHS satisfies the \textit{reproducing property}, namely, ${\left\langle {f\left| {\kappa (\mathbf{u},.)} \right.} \right\rangle _{{\mathcal{H}_k}}} = f(\mathbf{u})$, (2) can be rewritten as
\begin{equation}
\mathop {\min }\limits_{f \in {\mathcal{H}_k}} \sum\limits_{i = 1}^N {{{\left( {y(i) - {{\left\langle {f\left| {\kappa (\mathbf{u}(i),.)} \right.} \right\rangle }_{{\mathcal{H}_k}}}} \right)}^2}}  + \gamma \left\| f \right\|_{{\mathcal{H}_k}}^2
\end{equation}
By the \textit {representer theorem} \cite{burges1998tutorial}, the solution of (2) can be expressed as a linear combination of kernels:
\begin{equation}
f(\mathbf{u}) = \sum\limits_{i = 1}^N {{\alpha _i}\kappa \left( {\mathbf{u}(i),\mathbf{u}} \right)}
\end{equation}
The coefficient vector can be calculated as $\mathbf{\alpha} = {\left( {\mathbf{K} + \gamma \mathbf{I}} \right)^{ - 1}}\mathbf{y}$, where $\mathbf{K}{ \in \mathbb{R}^{N \times N}}$ is the Gram matrix with elements ${\mathbf{K}_{ij}} = \kappa \left( {\mathbf{u}(i),\mathbf{u}(j)} \right)$, and $\mathbf{y} = {[ {y(1), \cdots ,y(N)}]^T}$.

Solving the previous least squares problem usually requires significant memory and computational burden due to the necessity of calculating a large Gram matrix, whose dimension equals the number of input patterns. The KAF algorithms, however, provide efficient alternatives that build the solution incrementally, without explicitly computing the Gram matrix. Denote $f_i$ the estimated mapping (hypothesis) at iteration $i$. The KLMS algorithm can be expressed as \cite{principe2011kernel}
\begin{equation}
\left\{ \begin{array}{l}
{f_0} = 0\\
{f_i} = {f_{i - 1}} + \eta \kappa \left( {\mathbf{u}(i),.} \right)e(i)
\end{array} \right.{\rm{  }}
\end{equation}
where $\eta $ denotes the step size, $e(i)$ is the instantenous prediction error at iteration $i$, $e(i) = \mathbf{y}(i) - {f_{i - 1}}\left( {\mathbf{u}(i)} \right)$ , i.e. the instantenous error only depends upon the difference between the desired response at the current time and the evaluation of the current sample ($\mathbf{u}(i)$) with the previous system model ($f_{i - 1}$). The learned mapping of KLMS, at iteration $N$, will be
\begin{equation}
{f_N}(\mathbf{u}) = \eta \sum\limits_{i = 1}^N {e(i)\kappa \left( {\mathbf{u}(i),\mathbf{u}} \right)}
\end{equation}
This is a very nice result because it states that the solution to the unknown nonlinear mapping is done incrementally one step at a time, with a growing RBF network, where the centers are the samples and the fitting parameter is automatically determined as the current error.

Taking advantage of the incremental nature of the KLMS updates, the KLMS adaptation is in essence the solution of the following incremental regularized least squares problem:
\begin{equation}
\mathop {\min }\limits_{{f_i} \in {\mathcal{H}_k}} {\left( {y(i) - {f_i}\left( {\mathbf{u}(i)} \right)} \right)^2} + \frac{{1 - \eta }}{\eta }\left\| {{f_i} - {f_{i - 1}}} \right\|_{{\mathcal{H}_k}}^2
\end{equation}
Letting $\Delta {f_i} = {f_i} - {f_{i - 1}}$, (7) is equivalent to
\begin{equation}
\mathop {\min }\limits_{\Delta {f_i} \in {\mathcal{H}_k}} {\left( {e(i) - \Delta {f_i}\left( {\mathbf{u}(i)} \right)} \right)^2} + \frac{{1 - \eta }}{\eta }\left\| {\Delta {f_i}} \right\|_{{\mathcal{H}_k}}^2
\end{equation}

From (8), one may observe: 1) KLMS learning at iteration $i$ is equivalent to solving a regularized least squares problem, in which the previous hypothesis $f_{i - 1}$ is frozen, and only the adjustment term $\Delta {f_i}$ is optimized; 2) in this least squares problem, there is only one training example involved, i.e.$\left\{ {\mathbf{u}(i),e(i)} \right\}$; 3) the regularization factor is directly related to the step-size via $\gamma  = {{(1 - \eta )} \mathord{\left/{\vphantom {{(1 - \eta )} \eta }} \right. \kern-\nulldelimiterspace} \eta }$.

In the rest of the paper, the Mercer kernel $\kappa \left( {\mathbf{u},\mathbf{u}'} \right)$ is assumed to be the Gaussian kernel. In addition, to explicitly show the kernel size dependence, we denote the Gaussian kernel by $\kappa_\sigma \left( {\mathbf{u},\mathbf{u}'} \right)$, and the induced RKHS by $\mathcal{H}_\sigma$.

\section{KLMS with Adaptive Kernel Size}

The kernel is a crucial factor in all kernel methods in the sense that it defines the similarity between data points. For the Gaussian kernel, this similarity depends on the kernel size selected. If the kernel size is too large, then all the data will look similar in the RKHS (with inner products all close to 1), and the procedure reduces to linear regression. On the other hand, if the kernel size is too small, then all the data will look distinct (with inner products all close to 0), and the system fails to do inference on unseen data that fall between the training points.

Up to now the KLMS has been only studied with a constant kernel size, so all the elegance of the solution has not been fully recognized. In fact, the sequential learning algorithm (5) builds the current estimate of $f$ from two additive parts: the previous hypothesis $f_{i-1}$ and a correction term proportional to the prediction error on new data. In principle we can possibly use one RKHS to compute the previous hypothesis and change the RKHS to compute the correction term in (5), which can be efficiently done by changing the kernel size. This is the motivating idea that we pursue in this paper, and it has two fundamental components: (1) we have to formalize this approach; (2) we have to find an easy way of implementing it from samples. In the following, we propose an approach to sequentially optimize the KLMS with a variable kernel size.

\subsection{Sequential Optimization of the Kernel Size in KLMS}

In order to determine an optimal kernel size for KLMS, one should define in advance a cost function for the optimality. To make this precise, we suppose the training data $\{\mathbf{u}(i),y(i)\}$ are random, and there is an absolutely continuous probability measure $\mathit{P}$ (usually unknown) on the product space $\mathbb{U} \times \mathbb{Y}$ from which the data are drawn. The measure $\mathit{P}$ defines a regression function:
\begin{equation}
{f^*}(\mathbf{u}) = \int_{\mathbb{Y}} {yd\mathit{P}\left( {y\left| \mathbf{u} \right.} \right)}
\end{equation}
where $\mathit{P}\left( {y\left| \mathbf{u} \right.} \right)$ is the conditional measure on $\mathbf{u} \times \mathbb{Y}$. In this situation, the function $f^*$ can be said to be the desired mapping that needs to be estimated. Thus a measurement of the error in $f_i$  (which is updated by KLMS) is
\begin{equation}
{\mathcal{J}_1} = \int_{\mathbb{U}} {{{\left( {{f^*} - {f_i}} \right)}^2}dP(\mathbf{u})}
\end{equation}
where $\mathit{P}(\mathbf{u})$ is the marginal measure on $\mathbb{U}$. Then the  optimization should find the kernel size that minimizes this error. Since in practice $f^*$ is usually unknown, one can use the mean square error as an alternative cost for optimization:
\begin{equation}
{\mathcal{J}_2} = \int_{\mathbb{U} \times \mathbb{Y}} {{{\left[ {y - {f_i(\mathbf{u})}} \right]}^2}dP(\mathbf{u},y)}
\end{equation}
The cost $\mathcal{J}_2$ can be easily estimated from sample data. This is especially important when the probability measure $\mathit{P}$ is unknown.

Of course, the kernel size in KLMS can be optimized in batch mode, that is, the optimization is performed only after presenting the whole training data. Then, combining (6) and (11) yields the optimization:
\begin{equation}
{\sigma ^*} = \mathop {\arg \min }\limits_{\sigma { \in \mathbb{R} _ + }} \int_{\mathbb{U} \times \mathbb{Y}} {{{\left[ {y - \eta \sum\limits_{i = 1}^N {e(i){\kappa _\sigma }\left( {\mathbf{u}(i),\mathbf{u}} \right)} } \right]}^2}dP(\mathbf{u},y)}
\end{equation}
where $\sigma^{*}$ stands for the optimal kernel size. As KLMS is an online learning algorithm, we are more interested in a sequential optimization framework, which allows the kernel size to be sequentially optimized across iterations. To this end, we propose the following sequential optimization:
\begin{equation}
\sigma _i^* = \mathop {\arg \min }\limits_{{\sigma _i}{ \in \mathbb{R}_ + }} \int_{\mathbb{U} \times \mathbb{Y}} {{{\left[ {y - {f_{i - 1}}(\mathbf{u}) - \eta e(i){\kappa _{{\sigma _i}}}\left( {\mathbf{u}(i),\mathbf{u}} \right)} \right]}^2}dP(\mathbf{u},y)}
\end{equation}
where the previous hypothesis $f_{i-1}$ is frozen, and $\sigma_i$ denotes the kernel size at iteration $i$.

\textbf{\textit{Remark 1}}: By (13), the kernel size is optimized sequentially. Thus at iteration $i$, the initial learning step will determine an optimal value of the kernel size $\sigma_i$ (the old kernel sizes remain unchanged), followed by the addition of a new center using KLMS with this new kernel size. Learning with a varying kernel size implies at each iteration cycle to perform adaptation in a different RKHS since changing the kernel size modifies the inner product of the Hilbert space. For the KLMS, this learning paradigm is indeed feasible, because at each iteration cycle, the old centers remain frozen, and only a new center is added, or in other words, the correction term $\Delta {f_i}$ is just a feature vector in the current RKHS.

\textbf{\textit{Remark 2}}:A more reasonable approach should be to jointly optimize the kernel size and the step size (corresponding to the regularization factor). In this work, however, for simplicity the step size is assumed to be fixed and only the kernel size is optimized.

Suppose the training data $\{\mathbf{u}(i),y(i)\}$ are independent, identically distributed (i.i.d.). The hypothesis $f_i$, which depends on the previous training data, will be independent of the future training data. Then the mean square prediction error at iteration $i+1$, conditioned on $f_i$, equals
\begin{equation}
\begin{array}{l}
E\left[ {{e^2}(i + 1)\left| {{f_i}} \right.} \right]\\
 = \int_{\mathbb{U} \times \mathbb{Y}} {{{\left[ {y(i + 1) - {f_i}\left( {\mathbf{u}(i + 1)} \right)} \right]}^2}d\mathit{P}\left( {\mathbf{u}(i + 1),y(i + 1)\left| {{f_i}} \right.} \right)} \\
 = \int_{\mathbb{U} \times \mathbb{Y}} {{{\left[ {y(i + 1) - {f_i}\left( {\mathbf{u}(i + 1)} \right)} \right]}^2}d\mathit{P}\left( {\mathbf{u}(i + 1),y(i + 1)} \right)} \\
 = \int_{\mathbb{U} \times \mathbb{Y}} {{{\left[ {y - {f_i}\left( \mathbf{u} \right)} \right]}^2}d\mathit{P}(\mathbf{u},y)}
\end{array}
\end{equation}
where $\mathit{P}\left( \mathbf{u}(i + 1),y(i + 1)|f_i \right)$ denotes the probability measure of $\left( \mathbf{u}(i + 1),y(i + 1) \right)$ conditioned on $f_i$. The mapping update at iteration $i$ will affect directly the prediction error at iteration $i+1$, according to (14) the sequential optimization problem (13) can then be equivalently defined to search a value of the kernel size $\sigma_i$ such that the conditional mean square error $E[e^2(i+1)|f_i]$ is minimized:
\begin{equation}
\begin{array}{l}
\sigma _i^* = \mathop {\arg \min }\limits_{{\sigma _i}{ \in \mathbb{R} _ + }} E\left[ {{e^2}(i + 1)\left| {{f_i}} \right.} \right]\\
{\rm{    }} = \mathop {\arg \min }\limits_{{\sigma _i}{ \in \mathbb{R} _ + }} E\left[ {{e^2}(i + 1)\left| {{f_{i - 1}} + \eta e(i){\kappa _{{\sigma _i}}}\left( {\mathbf{u}(i),.} \right)} \right.} \right]
\end{array}
\end{equation}
To understand the above optimization in more detail, we consider the nonlinear regression model in which the output data ${y(i)}$ are related to the input vectors $\{\mathbf{u}(i)\}$ via
\begin{equation}
y(i) = {f^*}\left( {\mathbf{u}(i)} \right) + v(i)
\end{equation}
where $f^*(.)$ denotes the unknown nonlinear mapping that needs to be estimated, and $v(i)$ stands for the disturbance noise. In this case, the prediction error $e(i)$ can be expressed as
\begin{equation}
e(i) = y(i) - {f_{i - 1}}(\mathbf{u}(i)) = {\tilde f_{i - 1}}(\mathbf{u}(i)) + v(i)
\end{equation}
where $\tilde f_{i - 1} \triangleq f^*-f_{i-1}$ is the residual mapping at iteration $i-1$. The mean square error at iteration $i+1$, conditioned on $f_i$, is
\begin{equation}
\begin{array}{l}
E\left[ {{e^2}(i + 1)\left| {{f_i}} \right.} \right]\\
 = \int\limits_{\mathbb{U} \times \mathbb{V}} {{{\left( {{{\tilde f}_i}(\mathbf{u}(i + 1)) + v(i + 1)} \right)}^2}d{P_{\mathbf{u}v}}\left( {i + 1} \right)} \\
 = \int\limits_{\mathbb{U} \times \mathbb{V}} {{{\left( \begin{array}{l}
{{\tilde f}_{i - 1}}(\mathbf{u}(i + 1)) + v(i + 1) - \\
\eta e(i){\kappa _{{\sigma _i}}}\left( {\mathbf{u}(i),\mathbf{u}(i + 1)} \right)
\end{array} \right)}^2}d{P_{\mathbf{u}v}}\left( {i + 1} \right)}
\end{array}
\end{equation}%
where $\mathbb{V} \in \mathbb{R}$ denotes the noise space, and $P_{\mathbf{u}v}\left( {i + 1} \right)$ denotes the probability measure of $\left(\mathbf{u}(i + 1) + v(i + 1)\right)$.

\textbf{\textit{Remark 3}}: One can see from (18) that the optimal kernel size at iteration $i$ depends upon the residual mapping $\tilde f_{i - 1}$, prediction error $e(i)$, step size $\eta$, and the joint distribution $P_{\mathbf{u}v}$, which is much different from the optimal kernel sizes in problems of density estimation. Theoretically, given the desired mapping and the joint distribution $P_{\mathbf{u}v}$, the  optimal kernel sizes can be solved sequentially. This is, however, a rather tedious and impractical procedure since we have to solve an involved nonlinear optimization at each iteration cycle. More importantly, in practice the desired mapping, the noise, and the input distribution are usually unknown. Next, we will develop a stochastic gradient algorithm to adapt the kernel size, without resorting to any prior knowledge.

\subsection{KLMS with Adaptive Kernel Size}
As discussed previously, at each iteration cycle, the kernel size in KLMS can be optimized by minimizing the mean square error at next iteration (conditioned on the learned mapping at the current iteration). In this sense, one can optimize the previous kernel size using the current prediction error; that is, at iteration $i$, when prediction error $e(i)$ is available, the kernel size $\sigma_{i-1}$ can be optimized.

Actually, the kernel size $\sigma_{i-1}$ can be simply optimized by minimizing the instantaneous square error at iteration $i$, and a stochastic gradient algorithm can be readily derived as follows:
\begin{align} \nonumber
{{\sigma '}_{i - 1}} &= {\sigma _{i - 1}} - \mu \frac{\partial }{{\partial {\sigma _{i - 1}}}}\left[ {{e^2}(i)} \right]\\ \nonumber
{\rm{       }} &= {\sigma _{i - 1}} - 2\mu e(i)\frac{\partial }{{\partial {\sigma _{i - 1}}}}\left[ {{{\tilde f}_{i - 1}}(\mathbf{u}(i)) + v(i)} \right]\\
{\rm{       }} &= {\sigma _{i - 1}} - 2\mu e(i)\frac{\partial }{{\partial {\sigma _{i - 1}}}}\left[ \begin{array}{l}
{{\tilde f}_{i - 2}}(\mathbf{u}(i)) + v(i)\\
{\rm{  }} - \eta e(i - 1) \times \\
{\kappa _{{\sigma _{i - 1}}}}\left( {\mathbf{u}(i - 1),\mathbf{u}(i)} \right)
\end{array} \right]\\ \nonumber
{\rm{       }}& \mathop  = \limits^{(a)} {\sigma _{i - 1}} + \rho e(i - 1)e(i)\frac{\partial }{{\partial {\sigma _{i - 1}}}}{\kappa _{{\sigma _{i - 1}}}}\left( {\mathbf{u}(i - 1),\mathbf{u}(i)} \right)\\ \nonumber
{\rm{       }} &= {\sigma _{i - 1}} + \left( \begin{array}{l}
\rho e(i - 1)e(i){\left\| {\mathbf{u}(i - 1) - \mathbf{u}(i)} \right\|^2} \times \\
{\rm{       }}{{{\kappa _{{\sigma _{i - 1}}}}\left( {\mathbf{u}(i - 1),\mathbf{u}(i)} \right)} \mathord{\left/
 {\vphantom {{{\kappa _{{\sigma _{i - 1}}}}\left( {\mathbf{u}(i - 1),\mathbf{u}(i)} \right)} {\sigma _{i - 1}^3}}} \right.
 \kern-\nulldelimiterspace} {\sigma _{i - 1}^3}}
\end{array} \right)
\end{align}
where $\sigma^{'}_{i-1}$ denotes the updated kernel size at iteration $i-1$, (a) follows from the fact that $\tilde f_{i - 2}$ does not depend on $\sigma_{i-1}$, $\rho  = 2\mu \eta$ is the step size for the kernel size adaptation. At iteration $i$, however, the residual mapping ${\tilde f_{i - 1}}(u)$ has been frozen, and actually, the kernel size $\sigma_{i-1}$ cannot be modified. In this case, we just set $\sigma_{i}=\sigma^{'}_{i-1}$, and obtain the following sequential update algorithm:
%
\begin{equation}
\begin{array}{l}
{\sigma _i} = {\sigma _{i - 1}} + \rho e(i - 1)e(i) \times \\
{\rm{        }}{{{{\left\| {\mathbf{u}(i - 1) - \mathbf{u}(i)} \right\|}^2}{\kappa _{{\sigma _{i - 1}}}}\left( {\mathbf{u}(i - 1),\mathbf{u}(i)} \right)} \mathord{\left/
 {\vphantom {{{{\left\| {u(i - 1) - u(i)} \right\|}^2}{\kappa _{{\sigma _{i - 1}}}}\left( {u(i - 1),u(i)} \right)} {\sigma _{i - 1}^3}}} \right.
 \kern-\nulldelimiterspace} {\sigma _{i - 1}^3}}
\end{array}
\end{equation}

\textbf{\textit{Remark 4}}:The above algorithm is computationally very simple, since the kernel size is updated sequentially, where only the kernel size of the new center is updated, and those of the old centers remain frozen. The initial value of the kernel size can be set manually or calculated roughly using Silverman¡¯s rule based on the input distribution in advance.

From (20), we have the following observations:
\begin{description}
\item[1)]The direction of the gradient depends upon the signs of the prediction errors $e(i-1)$ and $e(i)$. Specifically, when the signs of $e(i-1)$ and $e(i)$ are the same, the kernel size will increase; while when the signs of $e(i-1)$ and $e(i)$ are different, the kernel size will decrease. This is reasonable, since in general the signs of two successive errors contain the information about the smoothness of the desired mapping. If there is little sign change, the desired mapping is likely a \textit{``moothing function''} and a larger kernel size is desirable; while if the sign changes frequently, the desired mapping is likely a\textit{ ``zig zag function''}, and in this case a smaller kernel size is usually better.
\item[2)]The magnitude of the gradient depends on the input data through \\
${\left\| {\mathbf{u}(i - 1) - \mathbf{u}(i)} \right\|^2}{\kappa _{{\sigma _{i - 1}}}}\left( {\mathbf{u}(i - 1),\mathbf{u}(i)}\right)$.The value of this term will be nearly zero when the distance between $\mathbf{u}(i - 1)$ and $\mathbf{u}(i)$ is very small or very large. This is also reasonable, since when $\mathbf{u}(i)$ is very close to $\mathbf{u}(i - 1)$, the sign change between $e(i-1)$
and $e(i)$ only implies a \textit{``very local fluctuation''}; while when $\mathbf{u}(i)$  is very far away from $\mathbf{u}(i - 1)$, the sign change between $e(i-1)$ and $e(i)$ contains little information about the smoothness of the desired mapping.
\item[3)]The magnitude of the gradient depends on $\sigma_{i-1}$ through\\ ${\kappa _{{\sigma _{i - 1}}}}\left( {\mathbf{u}(i - 1),\mathbf{u}(i)}\right)/\sigma^3_{i-1}$. For the case $\mathbf{u}(i-1)\neq\mathbf{u}(i)$,this term will approach zero when $\sigma_{i-1}$ is very small or very large. Therefore, the kernel size will be properly adjusted within a reasonable range.\\
    Combining (5) and (20), we obtain the KLMS with adaptive kernel size:
\begin{equation}
\left\{ \begin{array}{ll}
{f_0} =& 0\\
e(i) =& y(i) - {f_{i - 1}}(\mathbf{u}(i))\\
{\sigma _i} =& {\sigma _{i - 1}} + \rho e(i - 1)e(i){\left\| {\mathbf{u}(i - 1) - \mathbf{u}(i)} \right\|^2} \times \\
&{\rm{                    }}{{{\kappa _{{\sigma _{i - 1}}}}\left( {\mathbf{u}(i - 1),\mathbf{u}(i)} \right)} \mathord{\left/
 {\vphantom {{{\kappa _{{\sigma _{i - 1}}}}\left( {\mathbf{u}(i - 1),\mathbf{u}(i)} \right)} {\sigma _{i - 1}^3}}} \right.
 \kern-\nulldelimiterspace} {\sigma _{i - 1}^3}}\\
{f_i} =& {f_{i - 1}} + \eta e(i){\kappa _{{\sigma _i}}}\left( {\mathbf{u}(i),.} \right)
\end{array} \right.
\end{equation}
\end{description}

\textbf{\textit{Remark 5}}:The computational complexity of the algorithm (21) is in the same order of magnitude as that of the original KLMS algorithm, which equals $O(i)$ at iteration $i$.  This is because both algorithms share the same most time-consuming part, that is, the calculation of the prediction error.\\
Similar to the original KLMS algorithm, the new algorithm also produces a growing RBF network, whose network size increases linearly with the number of training data. In order to obtain a compact model (a network with as few centers as possible) and reduce the computational and memory costs, some sparsification or quantization methods can still be applied. Here, we only discuss the quantization approach. In \cite{chen2012quantized}, we use the idea of quantization to compress the input space of KLMS and constrain efficiently the network size growth. The learning rule of the quantized KLMS (QKLMS) is
\begin{equation}
\left\{ \begin{array}{l}
{f_0} = 0\\
e(i) = y(i) - {f_{i - 1}}(\mathbf{u}(i))\\
{f_i} = {f_{i - 1}} + \eta e(i){\kappa _\sigma }\left( {Q\left[ {\mathbf{u}(i)} \right],.} \right)
\end{array} \right.
\end{equation}
where $Q[.]$ denotes a quantization operator in input space $\mathbb{U}$.  In QKLMS (22), the centers are limited to the quantization codebook $\textbf{\emph{C}}$, and the network size can never be larger than the size of the codebook $\textit{size}(\textbf{\emph{C}})$.
 At iteration $i$, we just add $\eta e(i)$ to the coefficient of the code-vector closest to the current input $\mathbf{u}(i)$. A simple online vector quantization (VQ) method was also proposed in \cite{chen2012quantized}.

 Now suppose the online VQ method is adopted and the kernel size of QKLMS is varying as a function of the codebook index. Denote $\varepsilon$ the quantization size, and $d(\mathbf{u}(i),\textbf{\emph{C}})$ the distance between $\mathbf{u}(i)$ and $\textbf{\emph{C}}$:
$d\left( {\mathbf{u}(i),\textbf{\emph{C}}} \right) = {\min _{1 \le j \le size\left( \emph{C}\right)}}\left\| {\mathbf{u}(i) - {\textbf{\emph{C}}_j}} \right\|$, where $\textbf{\emph{C}}_j$ denotes the \textit{j}th element of the codebook $\textbf{\emph{C}}$.
Then at iteration \textit{i}, if $d(\mathbf{u}(i),\textbf{\emph{C}})>\varepsilon$, a new code-vector $(\mathbf{u}(i))$ will be added into the codebook, i.e. $\textbf{\emph{C}}=\{\textbf{\emph{C}},\mathbf{u}(i)\}$. In this case, a new center $(\mathbf{u}(i))$ will also be allocated, and its kernel size can be computed in a similar way as in (20):
\begin{equation}
{\sigma _j} = {\sigma _{j - 1}} + \rho e(j - 1)e(j){\left\| {{\textbf{\emph{C}}_j} - {\textbf{\emph{C}}_{j - 1}}} \right\|^2}{{{\kappa _{{\sigma _{j - 1}}}}\left( {{\emph{\textbf{C}}_j},{\textbf{\emph{C}}_{j - 1}}} \right)} \mathord{\left/
 {\vphantom {{{\kappa _{{\sigma _{j - 1}}}}\left( {{\textbf{\emph{C}}_j},{\textbf{\emph{C}}_{j - 1}}} \right)} {\sigma _{j - 1}^3}}} \right.
 \kern-\nulldelimiterspace} {\sigma _{j - 1}^3}}
 \end{equation}
 where $\sigma_j$ denotes the kernel size corresponding to the \textit{j}th code-vector $\left(\textbf{\emph{C}}_j\right),\textit{j}=\textit{size}(\emph{\textbf{C}})$, and $e(j)$ denotes the prediction error at the iteration when the code-vector $\emph{\textbf{C}}_j$ is added. If $d(\mathbf{u}(i),\textbf{\emph{C}})\leq\varepsilon$, , there is no new center added and no kernel size update.

\section{Convergence Analysis}

This section gives theoretical results on convergence of the algorithm (21). The unknown system is assumed to be the nonlinear regression model given in (16).  First, let us define the \emph{a priori} error $e_a(i)$ and \textit{a posteriori} error $e_p(i)$
as follows:
\begin{equation}
{e_a}(i) = {\tilde f_{i - 1}}(\textbf{u}(i)),\quad {\rm{   }}{e_p}(i) = {\tilde f_i}(\textbf{u}(i))
\end{equation}
where $\tilde f_{i - 1}$ and $\tilde f_{i}$ are the residual mappings at iteration $i-1$ and $i$, respectively.

\subsection{Kernel Size Convergence}

The exact analysis of the convergence of kernel size $\sigma_i$ is complex. In the following, we only show under several assumptions that the difference between two successive kernel sizes will converge in the mean to zero. These assumptions are:

\textbf{A1:} The noise $\nu(i)$ is zero-mean, independent, identically distributed, and independent of the input sequence $\{\textbf{u}(i)\}$;

\textbf{A2:} The step-sizes $\eta$ and $\rho$ are relative small such that as $i \to \infty $, the prediction errors $e(i)$ and $e(i-1)$ are independent of $\left\| {\textbf{u}(i - 1) -\textbf{ u}(i)} \right\|$ and the kernel size $\sigma_{i-1}$;

\textbf{A3:} As $i \to \infty $, the \textit{a priori} errors $e_a(i)$ and $e_a(i-1)$ are zero-mean and uncorrelated.

	Under assumptions \textbf{A1}-\textbf{A3}, we have, as $i \to \infty $,
\begin{equation}
\begin{array}{l}
E\left[ {{\sigma _i}} \right] - E\left[ {{\sigma _{i - 1}}} \right]\\
 = E\left[ {{{\rho e(i - 1)e(i){{\left\| {\textbf{u}(i - 1) - \textbf{u}(i)} \right\|}^2}{\kappa _{{\sigma _{i - 1}}}}\left( {\textbf{u}(i - 1),\textbf{u}(i)} \right)} \mathord{\left/
 {\vphantom {{\rho e(i - 1)e(i){{\left\| {\textbf{u}(i - 1) - \textbf{u}(i)} \right\|}^2}{\kappa _{{\sigma _{i - 1}}}}\left( {\textbf{u}(i - 1),\textbf{u}(i)} \right)} {\sigma _{i - 1}^3}}} \right.
 \kern-\nulldelimiterspace} {\sigma _{i - 1}^3}}} \right]\\
\mathop  = \limits^{A2} \rho E\left[ {e(i - 1)e(i)} \right]E\left[ {{{{{\left\| {\textbf{u}(i - 1) - \textbf{u}(i)} \right\|}^2}{\kappa _{{\sigma _{i - 1}}}}\left( {\textbf{u}(i - 1),\textbf{u}(i)} \right)} \mathord{\left/
 {\vphantom {{{{\left\| {\textbf{u}(i - 1) - \textbf{u}(i)} \right\|}^2}{\kappa _{{\sigma _{i - 1}}}}\left( {\textbf{u}(i - 1),\textbf{u}(i)} \right)} {\sigma _{i - 1}^3}}} \right.
 \kern-\nulldelimiterspace} {\sigma _{i - 1}^3}}} \right]\\
\mathop  = \limits^{A1} \rho E\left[ {{e_a}(i - 1){e_a}(i)} \right]E\left[ {{{{{\left\| {\textbf{u}(i - 1) - \textbf{u}(i)} \right\|}^2}{\kappa _{{\sigma _{i - 1}}}}\left( {\textbf{u}(i - 1),\textbf{u}(i)} \right)} \mathord{\left/
 {\vphantom {{{{\left\| {\textbf{u}(i - 1) - \textbf{u}(i)} \right\|}^2}{\kappa _{{\sigma _{i - 1}}}}\left( {\textbf{u}(i - 1),\textbf{u}(i)} \right)} {\sigma _{i - 1}^3}}} \right.
 \kern-\nulldelimiterspace} {\sigma _{i - 1}^3}}} \right]\\
\mathop  = \limits^{A3} 0
\end{array}
\end{equation}
which does not in itself imply convergence, but implies the difference between two successive kernel sizes will converge in mean to zero.

\textbf{\textit{Remark 6:}} The assumption \textbf{A1} is commonly used in the convergence analysis for adaptive filtering algorithms \cite{sayed2003fundamentals}. This assumption implies the independence between $\nu(i)$ and $e_a(i)$. The assumption \textbf{A2} is reasonable, since when the step size $\eta$ is very small, the steady-state misadjustment will be much smaller than the noise variance. In this case we have, as $i \to \infty $,$e(i - 1) \approx v(i - 1)$ and $e(i) \approx v(i)$. And hence, by the assumption \textbf{A1},  $e(i)$ and $e(i-1)$ are approximately independent of $\left\| {\textbf{u}(i - 1) - \textbf{u}(i)} \right\|$ . Further, if the step-size  $rho$ is also small, $e(i)$ and $e(i-1)$ will be approximately independent of the kernel size $\sigma_{i-1}$. The assumption \textbf{A3} will be easily met if the input sequence $\{\textbf{u}(i)\}$ is i.i.d.

\subsection{Energy Conservation Relation}
In adaptive filtering theory, the energy conservation relation provides a powerful tool for the mean square convergence analysis \cite{sayed2003fundamentals,al1900adaptive,yousef2001unified,al2003transient,al2003transients}. In our recent studies \cite{chen2012quantized,zhao2011kernel,chen2012mean}, this important relation has been extended into the RKHS. Before carrying out the mean square convergence analysis for the mapping update in (21), we derive the corresponding energy conservation relation.

The mapping update in (21) can be expressed as the residual-mapping update:
\begin{equation}
{\tilde f_i} = {\tilde f_{i - 1}} - \eta e(i){\kappa _{{\sigma _i}}}\left( {\textbf{u}(i),.} \right)
\end{equation}

 Due to the variable kernel size, at each iteration the correction term in (26) is computed in a different RKHS. In order to derive the energy conservation relation in a fixed RKHS, one should find a RKHS that contains all the correction terms. Here we give an important lemma.

\textbf{\textit{Lemma 1: }}Let $\mathbb{U}\in \mathbb{R}^m$ be any set with nonempty interior. Then the RKHS ${\mathbf{\mathcal{H}}_{{\sigma _*}}}$ induced by the Gaussian kernel ${\kappa _{{\sigma _*}}}(\textbf{u},\textbf{u}')$ on $\mathbb{U}$ contains the function ${\kappa _{{\sigma _*}}}(\textbf{u}\mathbf{,.})$ if and only if $\sigma  > {{\sqrt 2 {\sigma _*}} \mathord{\left/
 {\vphantom {{\sqrt 2 {\sigma _*}} 2}} \right.
 \kern-\nulldelimiterspace} 2}$ . For such $\sigma$, the function ${\kappa _{{\sigma _*}}}(\textbf{u}\mathbf{,.})$ has norm given by:
\begin{equation}
{\left\| {{\kappa _\sigma }\left( {\textbf{u},.} \right)} \right\|_{{\mathcal{H}_{{\sigma _*}}}}} = {\left( {\frac{{{\sigma ^2}}}{{{\sigma _*}\sqrt {2{\sigma ^2} - \sigma _*^2} }}} \right)^{m/2}}
\end{equation}

\textbf{\textit{Remark 7:}} The above lemma is a direct consequence of the Theorem 2 in \cite{quang2012further}.For the case $\sigma  < {{\sqrt 2 {\sigma _*}} \mathord{\left/
 {\vphantom {{\sqrt 2 {\sigma _*}} 2}} \right.
 \kern-\nulldelimiterspace} 2}$ ., the Hilbert space $\mathbf{\mathcal{H}}_{\sigma_*}$ will not contain the function ${\kappa _{{\sigma }}}(\textbf{u}\mathbf{,.})$. We point out that in this case, the function ${\kappa _{{\sigma}}}(\textbf{u}\mathbf{,.})$ can still be arbitrarily "close" to a member of $\mathbf{\mathcal{H}}_{\sigma_*}$, because $\mathbf{\mathcal{H}}_{\sigma_*}$ is dense in the space of continuous functions on $\mathbb{U}$ provided $\mathbb{U}$ is compact.

 Now we select a fixed kernel size $\sigma_*$, satisfying ${\sigma _*} < \sqrt 2 {\sigma _{\min }}$, where ${\sigma _{\min }} = \min \left\{ {{\sigma _i}} \right\}$. By Lemma 1, the RKHS $\mathbf{\mathcal{H}}_{\sigma_*}$ will contain all the correction terms.

By the \textit{reproducing property} of the RKHS $\mathbf{\mathcal{H}}_{\sigma_*}$, the prediction error $e(i)$, \textit{a priori} error $e_a(i)$, and a \textit{posteriori} error $e_p(i)$ can be expressed as
\begin{equation}
\left\{ \begin{array}{l}
e(i) = {\left\langle {{{\tilde f}_{i - 1}}\left| {{\kappa _{{\sigma _*}}}(\textbf{u}(i),.)} \right.} \right\rangle _{{\mathcal{H}_{{\sigma _*}}}}} + v(i)\\
{e_a}(i) = {\left\langle {{{\tilde f}_{i - 1}}\left| {{\kappa _{{\sigma _*}}}(\textbf{u}(i),.)} \right.} \right\rangle _{{\mathcal{H}_{{\sigma _*}}}}}\\
{e_p}(i) = {\left\langle {{{\tilde f}_i}\left| {{\kappa _{{\sigma _*}}}(\textbf{u}(i),.)} \right.} \right\rangle _{{\mathcal{H}_{{\sigma _*}}}}}
\end{array} \right.
\end{equation}
Further, one can derive the relationship between $e_a(i)$ and $e_p(i)$:
\begin{equation}
\begin{array}{ll}
{e_p}(i) &= {e_a}(i) - \eta e(i){\kappa _{{\sigma _i}}}\left( {\mathbf{u}(i),\mathbf{u}(i)} \right)\\
{\rm{        }}& = {e_a}(i) - \eta e(i)
\end{array}
\end{equation}
Hence
\begin{equation}
{\tilde f_i} = {\tilde f_{i - 1}} + \left( {{e_p}(i) - {e_a}(i)} \right){\kappa _{{\sigma _i}}}(\textbf{u}(i),.)
\end{equation}
Squaring both sides of (30) in RKHS $\mathbf{\mathcal{H}}_{\sigma_*}$, we derive
\[
\begin{array}{lll}
\left\| {{{\tilde f}_i}} \right\|_{{\mathcal{H}_{{\sigma _*}}}}^2 &= &\left\| {{{\tilde f}_{i - 1}}} \right\|_{{\mathcal{H}_{{\sigma _*}}}}^2 + {\left( {{e_p}(i) - {e_a}(i)} \right)^2}{\left\langle {{\kappa _{{\sigma _i}}}(\mathbf{u}(i),.)\left| {{\kappa _{{\sigma _i}}}(\mathbf{u}(i),.)} \right.} \right\rangle _{{\mathcal{H}_{{\sigma _*}}}}}\\
{\rm{                }} &&+ 2\left( {{e_p}(i) - {e_a}(i)} \right){\left\langle {{{\tilde f}_{i - 1}}\left| {{\kappa _{{\sigma _i}}}(\mathbf{u}(i),.)} \right.} \right\rangle _{{\mathcal{H}_{{\sigma _*}}}}}\\
 &= &\left\| {{{\tilde f}_{i - 1}}} \right\|_{{\mathcal{H}_{{\sigma _*}}}}^2 + {\left( {{e_p}(i) - {e_a}(i)} \right)^2}{\left\langle {{\kappa _{{\sigma _i}}}(\mathbf{u}(i),.)\left| {{\kappa _{{\sigma _*}}}(\mathbf{u}(i),.) + {\bm{\delta} _i}(.)} \right.} \right\rangle _{{\mathcal{H}_{{\sigma _*}}}}}\\
{\rm{           }} &&+ 2\left( {{e_p}(i) - {e_a}(i)} \right){\left\langle {{{\tilde f}_{i - 1}}\left| {{\kappa _{{\sigma _*}}}(\mathbf{u}(i),.) + {\bm{\delta} _i}(.)} \right.} \right\rangle _{{\mathcal{H}_{{\sigma _*}}}}}\\
 &= &\left\| {{{\tilde f}_{i - 1}}} \right\|_{{\mathcal{H}_{{\sigma _*}}}}^2 + e_p^2(i) - e_a^2(i) + \varepsilon (i)
\end{array}
\]
where $\left\| {{{\tilde f}_i}} \right\|_{{\mathcal{H}_{{\sigma _*}}}}^2 = {\left\langle {{{\tilde f}_i}\left| {{{\tilde f}_i}} \right.} \right\rangle _{{\mathcal{H}_{{\sigma _*}}}}}$ is the residual mapping power (RMP) at iteration \textit{i}, ${\bm{\delta} _i}(.) = {\kappa _{{\sigma _i}}}(\mathbf{u}(i),.) - {\kappa _{{\sigma _*}}}(\mathbf{u}(i),.)$, and

\[\varepsilon (i) = {\left\langle {\left[ {{{\left( {{e_p}(i) - {e_a}(i)} \right)}^2}{\kappa _{{\bm{\sigma} _i}}}(\mathbf{u}(i),.) + 2\left( {{e_p}(i) - {e_a}(i)} \right){{\tilde f}_{i - 1}}} \right]\left| {{\bm{\delta} _i}(.)} \right.} \right\rangle _{{\mathcal{H}_{{\sigma _*}}}}}\]
It follows that
\begin{equation}
\left\| {{{\tilde f}_i}} \right\|_{{\mathcal{H}_{{\sigma _*}}}}^2 + e_a^2(i) = \left\| {{{\tilde f}_{i - 1}}} \right\|_{{\mathcal{H}_{{\sigma _*}}}}^2 + e_p^2(i) + \varepsilon (i)
\end{equation}

\textbf{\textit{Remark 8:}}  Equation (31) is referred to as the \textit{energy conservation relation} for KLMS with adaptive kernel size. If the kernel size is fixed, say ${\sigma _i} \equiv {\sigma _*}$, we have ${\delta _i}(.) = 0$, and hence $\varepsilon (i) = 0$. In this case, (31) reduces to the energy conservation relation for the original KLMS:
\begin{equation}
\left\| {{{\tilde f}_i}} \right\|_{{H_{{\sigma _*}}}}^2 + e_a^2(i) = \left\| {{{\tilde f}_{i - 1}}} \right\|_{{H_{{\sigma _*}}}}^2 + e_p^2(i)
\end{equation}
which, in form, is identical to the energy conservation relation for the normalized LMS (NLMS) algorithm.

\subsection{Sufficient Condition for Mean Square Convergence }
Substituting ${e_p}(i) = {e_a}(i) - \eta e(i)$ into (31) and taking expectations of both sides yield
\begin{equation}
\begin{array}{l}
E\left[ {\left\| {{{\tilde f}_i}} \right\|_{{\mathcal{H}_{{\sigma _*}}}}^2} \right] - E\left[ {\left\| {{{\tilde f}_{i - 1}}} \right\|_{{\mathcal{H}_{{\sigma _*}}}}^2} \right]\\
 =  - 2\eta E\left[ {e(i)\left( {{e_a}(i) + {{\left\langle {{{\tilde f}_{i - 1}}\left| {{\bm{\delta} _i}(.)} \right.} \right\rangle }_{{\mathcal{H}_{{\sigma _*}}}}}} \right)} \right]\\
{\rm{              }}\qquad + {\eta ^2}E\left[ {{e^2}(i)\left( {1 + {{\left\langle {{\kappa _{{\sigma _i}}}(\mathbf{u}(i),.)\left| {{\bm{\delta} _i}(.)} \right.} \right\rangle }_{{\mathcal{H}_{{\sigma _*}}}}}} \right)} \right]\\
\mathop  = \limits^{(b)} - 2\eta E\left[ {e(i)\left( {{e_a}(i) + {{\left\langle {{{\tilde f}_{i - 1}}\left| {{\bm{\delta} _i}(.)} \right.} \right\rangle }_{{\mathcal{H}_{{\sigma _*}}}}}} \right)} \right]\\
{\rm{              }} \qquad+ {\eta ^2}E\left[ {{e^2}(i)\left( {1 + \left\| {{\bm{\delta} _i}(.)} \right\|_{{\mathcal{H}_{{\sigma _*}}}}^2} \right)} \right]
\end{array}
\end{equation}
where (b) follows from
\begin{equation}
\begin{array}{l}
{\left\langle {{\kappa _{{\sigma _i}}}(\textbf{u}(i),.)\left| {{\bm{\delta }_i}(.)} \right.} \right\rangle _{{\mathcal{H}_{{\sigma _*}}}}} = {\left\langle {{\bm{\delta } _i}(.) + {\kappa _{{\sigma _*}}}(\mathbf{u}(i),.)\left| {{\bm{\delta } _i}(.)} \right.} \right\rangle _{{\mathcal{H}_{{\sigma _*}}}}}\\
{\rm{             }} = {\left\langle {{\kappa _{{\sigma _*}}}(\mathbf{u}(i),.)\left| {{\bm{\delta } _i}(.)} \right.} \right\rangle _{{\mathcal{H}_{{\sigma _*}}}}} + \left\| {{\bm{\delta } _i}(.)} \right\|_{{\mathcal{H}_{{\sigma _*}}}}^2\\
{\rm{             }} = {\bm{\delta } _i}\left( {\mathbf{u}(i)} \right) + \left\| {{\bm{\delta } _i}(.)} \right\|_{{\mathcal{H}_{{\sigma _*}}}}^2\\
{\rm{             }} = \left\| {{\bm{\delta } _i}(.)} \right\|_{{\mathcal{\mathcal{H}}_{{\sigma _*}}}}^2
\end{array}
\end{equation}
It follows easily that
\begin{equation}
\begin{array}{l}
E\left[ {\left\| {{{\tilde f}_i}} \right\|_{{\mathcal{H}_{{\sigma _*}}}}^2} \right] \le E\left[ {\left\| {{{\tilde f}_{i - 1}}} \right\|_{{\mathcal{H}_{{\sigma _*}}}}^2} \right]\\
{\rm{ }}\; \Leftrightarrow \eta  \le \frac{\displaystyle{{2E\left[ {e(i)\left( {{e_a}(i) + {{\left\langle {{{\tilde f}_{i - 1}}\left| {{\bm{\delta} _i}(.)} \right.} \right\rangle }_{{\mathcal{H}_{{\sigma _*}}}}}} \right)} \right]}}}{\displaystyle{{E\left[ {{e^2}(i)\left( {1 + \left\| {{\bm{\delta } _i}(.)} \right\|_{{\mathcal{H}_{{\sigma _*}}}}^2} \right)} \right]}}}
\end{array}
\end{equation}
Thus, if $\forall i$, the step-size $\eta$ satisfies the inequality
\begin{equation}
0 < \eta  \le \frac{\displaystyle{{2E\left[ {e(i)\left( {{e_a}(i) + {{\left\langle {{{\tilde f}_{i - 1}}\left| {{\bm{\delta} _i}(.)} \right.} \right\rangle }_{{\mathcal{H}_{{\sigma _*}}}}}} \right)} \right]}}}{\displaystyle{{E\left[ {{e^2}(i)\left( {1 + \left\| {{\bm{\delta} _i}(.)} \right\|_{{\mathcal{H}_{{\sigma _*}}}}^2} \right)} \right]}}}
\end{equation}
the RMP in RKHS $\mathbf{\mathcal{H}}_{\sigma_*}$  will monotonically decrease (and hence converge).  The inequality (36) implies
$\forall i$
\[E\left[ {e(i)\left( {{e_a}(i) + {{\left\langle {{{\tilde f}_{i - 1}}\left| {{\bm{\delta}_i}(.)} \right.} \right\rangle }_{{\mathcal{H}_{{\sigma _*}}}}}} \right)} \right] > 0\]
So a sufficient condition for the mean square convergence (monotonic decrease of the RMP) will be,$\forall i$,
\begin{equation}
\left\{ \begin{array}{l}
E\left[ {e(i)\left( {{e_a}(i) + {{\left\langle {{{\tilde f}_{i - 1}}\left| {{\bm{\delta} _i}(.)} \right.} \right\rangle }_{{H_{{\sigma _*}}}}}} \right)} \right] > 0\\
0 < \eta  \le \frac{\displaystyle{{2E\left[ {e(i)\left( {{e_a}(i) + {{\left\langle {{{\tilde f}_{i - 1}}\left| {{\bm{\delta} _i}(.)} \right.} \right\rangle }_{{H_{{\sigma _*}}}}}} \right)} \right]}}}{\displaystyle{{E\left[ {{e^2}(i)\left( {1 + \left\| {{\bm{\delta} _i}(.)} \right\|_{{H_{{\sigma _*}}}}^2} \right)} \right]}}}
\end{array} \right.{\rm{ }}
\end{equation}

\subsection{Steady-State Excess Mean Square Error}
Further, we take the limit of (33) as $i \to \infty $:
\begin{equation}
\begin{array}{l}
\mathop {\lim }\limits_{i \to \infty } E\left[ {\left\| {{{\tilde f}_i}} \right\|_{{H_{{\sigma _*}}}}^2} \right] - \mathop {\lim }\limits_{i \to \infty } E\left[ {\left\| {{{\tilde f}_{i - 1}}} \right\|_{{H_{{\sigma _*}}}}^2} \right]\\
 =  - 2\eta \mathop {\lim }\limits_{i \to \infty } E\left[ {e(i)\left( {{e_a}(i) + {{\left\langle {{{\tilde f}_{i - 1}}\left| {{\bm{\bm{\delta} } _i}(.)} \right.} \right\rangle }_{{H_{{\sigma _*}}}}}} \right)} \right]\\
{\rm{      }} + {\eta ^2}\mathop {\lim }\limits_{i \to \infty } E\left[ {{e^2}(i)\left( {1 + \left\| {{\bm{\bm{\delta} } _i}(.)} \right\|_{{\mathcal{H}_{{\sigma _*}}}}^2} \right)} \right]
\end{array}
\end{equation}
If the RMP reaches steady-state, that is
\begin{equation}
\mathop {\lim }\limits_{i \to \infty } E\left[ {\left\| {{{\tilde f}_i}} \right\|_{{\mathcal{H}_{{\sigma _*}}}}^2} \right] = \mathop {\lim }\limits_{i \to \infty } E\left[ {\left\| {{{\tilde f}_{i - 1}}} \right\|_{{\mathcal{H}_{{\sigma _*}}}}^2} \right]
\end{equation}
then the following relation holds:
\begin{equation}
\begin{array}{l}
\mathop {\lim }\limits_{i \to \infty } E\left[ {e(i)\left( {{e_a}(i) + {{\left\langle {{{\tilde f}_{i - 1}}\left| {{\bm{\delta}  _i}(.)} \right.} \right\rangle }_{{\mathcal{H}_{{\sigma _*}}}}}} \right)} \right]\\
{\rm{         }} = \frac{\displaystyle{\eta}}{\displaystyle{2}}\mathop {\lim }\limits_{i \to \infty } E\left[ {{e^2}(i)\left( {1 + \left\| {{\bm{\delta}  _i}(.)} \right\|_{{\mathcal{H}_{{\sigma _*}}}}^2} \right)} \right]
\end{array}
\end{equation}
In order to derive the steady-state excess mean-square error\footnote{The \textit{a priori} error power is also referred to as the excess mean-square error in literature of adaptive filtering.}(EMSE) $\mathop {\lim }\limits_{i \to \infty } E\left[ {e_a^2(i)} \right]$, we use two assumptions: one is the assumption \textbf{A1}, and another is as follows:
\textbf{A4:} The squared \textit{a priori} error $e_a^2(i)$ and $\left\| {{\bm{\delta}  _i}(.)} \right\|_{{\mathcal{H}_{{\sigma _*}}}}^2$ are uncorrelated \footnote{The assumption A4 will be easily met if the input sequence is i.i.d.}.

Under the assumptions \textbf{A1} and \textbf{A4}, (40) becomes
\begin{equation}
\begin{array}{l}
\mathop {\lim }\limits_{i \to \infty } E\left[ {e_a^2(i)} \right] + \mathop {\lim }\limits_{i \to \infty } E\left[ {{e_a}(i){{\left\langle {{{\tilde f}_{i - 1}}\left| {{\bm{\delta}  _i}(.)} \right.} \right\rangle }_{{\mathcal{H}_{{\sigma _*}}}}}} \right]\\
 = \frac{\displaystyle{\eta}}{\displaystyle{2}}\left( {\mathop {\lim }\limits_{i \to \infty } E\left[ {e_a^2(i)} \right] + \xi _v^2} \right)\left( {1 + \mathop {\lim }\limits_{i \to \infty } E\left[ {\left\| {{\bm{\delta}  _i}(.)} \right\|_{{\mathcal{H}_{{\sigma _*}}}}^2} \right]} \right)
\end{array}
\end{equation}
where $\xi _v^2$ is the noise variance. Then we have
\begin{equation}
\mathop {\lim }\limits_{i \to \infty } E\left[ {e_a^2(i)} \right] = \frac{{\eta \xi _v^2\left( {1 + \varsigma } \right) - 2\tau }}{{2 - \eta \left( {1 + \varsigma } \right)}}
\end{equation}
where $\varsigma  = \mathop {\lim }\limits_{i \to \infty } E\left[ {\left\| {{\bm{\delta}  _i}(.)} \right\|_{{\mathcal{H}_{{\sigma _*}}}}^2} \right]$, and $\tau  = \mathop {\lim }\limits_{i \to \infty } E\left[ {{e_a}(i){{\left\langle {{{\tilde f}_{i - 1}}\left| {{\bm{\delta}  _i}(.)} \right.} \right\rangle }_{{\mathcal{H}_{{\sigma _*}}}}}} \right]$. By Lemma 1, $\varsigma $ can also be expressed as
\begin{equation}
\begin{array}{ll}
\varsigma & = \mathop {\lim }\limits_{i \to \infty } E\left[ {\left\| {{\bm{\delta}  _i}(.)} \right\|_{{\mathcal{H}_{{\sigma _*}}}}^2} \right]\\
{\rm{  }} &= \mathop {\lim }\limits_{i \to \infty } E\left[ {\left\| {{\kappa _{{\sigma _i}}}(\mathbf{u}(i),.) - {\kappa _{{\sigma _*}}}(\mathbf{u}(i),.)} \right\|_{{\mathcal{H}_{{\sigma _*}}}}^2} \right]\\
{\rm{  }} &= \mathop {\lim }\limits_{i \to \infty } E\left[ {\left\| {{\kappa _{{\sigma _i}}}(\mathbf{u}(i),.)} \right\|_{{\mathcal{H}_{{\sigma _*}}}}^2} \right] - 1\\
{\rm{  }} &= \mathop {\lim }\limits_{i \to \infty } E\left[ {{{\left( {\frac{{\sigma _i^2}}{{{\sigma _*}\sqrt {2\sigma _i^2 - \sigma _*^2} }}} \right)}^m}} \right] - 1
\end{array}
\end{equation}

\textbf{\textit{Remark 9:}}Although both $\varsigma $ and $\tau$ depend on the kernel size $\sigma_*$, we should note that the steady-state EMSE itself does not depend on $\sigma_*$. This can be easily understood by the fact that the residual mapping $\widetilde{f}_i$ has no relation to $\sigma_*$.

To further investigate the steady-state EMSE, we consider the case in which as $i \to \infty $, $\sigma_i$ and ${\bar \sigma _\infty }$ are very close and satisfy $\left| {{\sigma _i} - {{\bar \sigma }_\infty }} \right| < {{\left( {2 - \sqrt 2 } \right){{\bar \sigma }_\infty }} \mathord{\left/
 {\vphantom {{\left( {2 - \sqrt 2 } \right){{\bar \sigma }_\infty }} 2}} \right.
 \kern-\nulldelimiterspace} 2}$, where ${\bar \sigma _\infty } = \mathop {\lim }\limits_{i \to \infty } E\left[ {{\sigma _i}} \right]$. In this case we have ${\bar \sigma _\infty } < \sqrt 2 {\sigma _i}$ as $i \to \infty $. Then at the steady-state stage, we can set . And hence
 \begin{equation}
\varsigma  = \mathop {\lim }\limits_{i \to \infty } E\left[ {{{\left( {\frac{{\sigma _i^2}}{{{\sigma _*}\sqrt {2\sigma _i^2 - \sigma _*^2} }}} \right)}^m} - 1} \right] \approx 0
 \end{equation}
 and
 \begin{equation}
\begin{array}{ll}
\left| \tau  \right| &= \mathop {\lim }\limits_{i \to \infty } \left| {E\left[ {{e_a}(i){{\left\langle {{{\tilde f}_{i - 1}}\left| {{\bm{\delta}  _i}(.)} \right.} \right\rangle }_{{\mathcal{H}_{{\sigma _*}}}}}} \right]} \right|\\
{\rm{  }} &\le \mathop {\lim }\limits_{i \to \infty } E\left[ {\left| {{e_a}(i){{\left\langle {{{\tilde f}_{i - 1}}\left| {{\bm{\delta}  _i}(.)} \right.} \right\rangle }_{{\mathcal{H}_{{\sigma _*}}}}}} \right|} \right]\\
{\rm{  }} &\le \mathop {\lim }\limits_{i \to \infty } E\left[ {\left| {{e_a}(i)} \right|{{\left\| {{{\tilde f}_{i - 1}}} \right\|}_{{\mathcal{H}_{{\sigma _*}}}}}{{\left\| {{\bm{\bm{\delta} } _i}(.)} \right\|}_{{\mathcal{H}_{{\sigma _*}}}}}} \right]\\
{\rm{  }} &\approx 0
\end{array}
 \end{equation}
It follows that
\begin{equation}
\mathop {\lim }\limits_{i \to \infty } E\left[ {e_a^2(i)} \right] = \frac{{\eta \xi _v^2\left( {1 + \varsigma } \right) - 2\tau }}{{2 - \eta \left( {1 + \varsigma } \right)}} \approx \frac{{\eta \xi _v^2}}{{2 - \eta }}
\end{equation}

\textbf{\textit{Remark 10}}: It has been shown that the steady-state EMSE of the original KLMS (with a fixed kernel size) equals ${{\eta \xi _v^2} \mathord{\left/{\vphantom {{\eta \xi _v^2} {(2 - \eta )}}} \right. \kern-\nulldelimiterspace} {(2 - \eta )}}$, which is not related to the specific value of the kernel size \cite{chen2012mean}. From (46) one observes that, when the kernel size $\sigma_i$ converges to a neighborhood of a certain constant (${\bar \sigma _\infty }$), the adaptation of kernel size also has little effect on the steady-state EMSE. This will be confirmed later by simulation results. We should point out here that, although the kernel size may have little effect on the KLMS steady-state accuracy (in terms of the EMSE), it has significant influence on the convergence speed. In most practical situations, the training data are finite and the algorithm can never reach the steady state. In these cases the kernel size also has significant influence on the final accuracy (not the steady-state accuracy).

\section{Simulation Results}
In this section, we present simulation results that illustrate the performance of the proposed algorithm. The simulation examples presented include static function approximation and short-term chaotic time series prediction.
\subsection{Static Function Approximation}
Consider a simple static function estimation problem in which the desired output data are generated by
\begin{equation}
y(i) = \cos (8u(i)) + v(i)
\end{equation}
where the input $u(i)$ is uniformly distributed over $[-\pi,\pi]$, and  is $\{v(i)\}$ a white Gaussian noise with variance 0.0001.

For the KLMS with different kernel sizes, the average convergence curves (in terms of the EMSE) over 1000 Monte Carlo runs are shown in Fig. 1. In the simulation, the step-sizes for all the cases are set at $\eta=0.5$.  For the KLMS with adaptive kernel size, the step-size for the kernel size adaptation is set at $\rho=0.025$, and the initial kernel size is set as 1.0. From Fig. 1, we see clearly that the kernel size has significant influence on the convergence speed. In this example, the kernel size $\sigma=1.0$ and $\sigma=0.5$ produce a rather slow convergence speed. The kernel sizes $\sigma=0.05$ and $\sigma=0.1$ work very well, and in particular, the kernel size $\sigma=0.1$ achieves a fast convergence speed and the smallest final EMSE (at the 5000th iteration). The kernel size $\sigma=0.35$ (selected by Silverman¡¯s rule) works, but obviously the performance is not so good. Although the initial kernel size is set as 1.0 (with which the algorithm is almost stalled), the KLMS with adaptive kernel size ($\sigma_i$) can still converge to a very small EMSE at the final iteration. This can be clearly explained from Fig. 2, where the evolution curve of the adaptive kernel size $\sigma_i$ has been plotted. In Fig. 2, the adaptive kernel size $\sigma_i$ converges to a desirable value between 0.1 and 0.2. Fig. 3 shows the learned mappings at final iteration for different kernel sizes. The desired mapping $f^{*}(u)=cos(8u)$ is also plotted in Fig. 3 for comparison purpose. One can see for the cases $\sigma=0.05$, $\sigma=0.05$ and $\sigma=\sigma_{i}$, the learned mappings match the desired mapping very well, while when $\sigma=1.0$ and 0.5, the learned mappings deviate severely from the desired function. For the kernel size $\sigma=0.35$, there is still some visible deviation between the learned mapping and the desired one. A more detailed comparison is also presented in Table 1, where the EMSE at final iteration is summarized.

As illustrated in Fig. 1, the initial convergence speed of the KLMS with adaptive kernel size can still be very slow if the initial kernel size is inappropriately chosen. In order to improve the initial convergence speed, one can select a suitable initial kernel size using a certain method such as Silverman¡¯s rule. For the present example, if we set the initial kernel size to be 0.35, the convergence speed of the new algorithm will be improved significantly. This can be clearly seen from Fig. 4, in which the learning curves for $\sigma=0.1$ and $\sigma=\sigma_i$ (with $\sigma_0=0.35$) are shown.

The kernel size will influence the convergence speed (see Fig. 1) and the final accuracy with finite training data (see Table 1), but it has little effect on the steady-state EMSE with infinite training data. In order to confirm this theoretical prediction, we perform another set of simulations with the same settings, except now much more iterations are run. For different kernel sizes and iterations, the EMSEs (obtained as the averages over a window of 2000 samples) are listed in Table 2. One can see for the kernel sizes $\sigma=0.05$, $\sigma=0.1$ and $\sigma=\sigma_i$, the algorithms almost reach the steady-state before the $100000^{th}$ iteration. For the kernel size $\sigma=0.35$, the algorithm attains its steady-state at around the $800000^{th}$ iteration. For the cases $\sigma=0.5$ and $\sigma=0.1$, it is hard to obtain the steady-state EMSE via simulation since the convergence speed is too slow. In Table 2, the simulated steady-state EMSEs for different kernel sizes ( $\sigma=0.05$, $\sigma=0.1$, $\sigma=\sigma_i$ and $\sigma=0.35$ ) are very close and approach to 0.000033333, the theoretical value of the steady-state EMSE calculated using (46).
\begin{table}
\begin{center}
\begin{tabular}{c|c}
\hline

\hline\noalign{\smallskip}
Kernel Size & EMSE at final iteration\\
\hline
$\sigma=0.05$ & $0.00006 \pm 0.00022$ \\
\hline
$\sigma=0.1$ & $0.00005 \pm 0.00010$ \\
\hline
$\sigma=0.35$ & $0.0019 \pm 0.0068$ \\
\hline
$\sigma=0.5$ & $0.3798 \pm 0.4468$ \\
\hline
$\sigma=1.0$ & $0.6573 \pm 0.6846$ \\
\hline
adaptive kernel size & $0.00007 \pm 0.00027$ \\
\hline

\hline
\noalign{\smallskip}
\end{tabular}
\caption{EMSE at final iteration for different kernel sizes}
\end{center}
\end{table}

\begin{table}
\begin{center}
\begin{tabular}{@{}c|c|c|c|c|c}
\hline

\hline
\diagbox {Kernel size} {Iterations} & 10000 & 50000 & 100000 & 200000	& 800000\\
\hline
$\sigma=0.05$ & 0.00004802 & 0.00003507 & 0.00003332 & 0.00003334 & 0.00003333 \\
\hline
$\sigma=0.1$ & 0.00003977 & 0.00003332 & 0.00003333 & 0.00003331 & 0.00003332 \\
\hline
$\sigma=0.35$ & 0.001042 & 0.0001523 & 0.00008837 & 0.00005514 & 0.00003334 \\
\hline
$\sigma=0.5$ & 0.2686 & 0.05134 & 0.01211 & 0.001312 & 0.0001208\\
\hline
$\sigma=1.0$ & 0.6457 & 0.6207 & 0.6024 & 0.5830 & 0.5684 \\
\hline
adaptive kernel size & 0.00004859 & 0.00003581 & 0.00003334 & 0.00003332 & 0.00003334 \\
\hline

\hline
\noalign{\smallskip}
\end{tabular}
\caption{EMSE at different iterations for different kernel sizes}
\end{center}
\end{table}

\begin{figure}
\centering
\includegraphics[height=5.5cm]{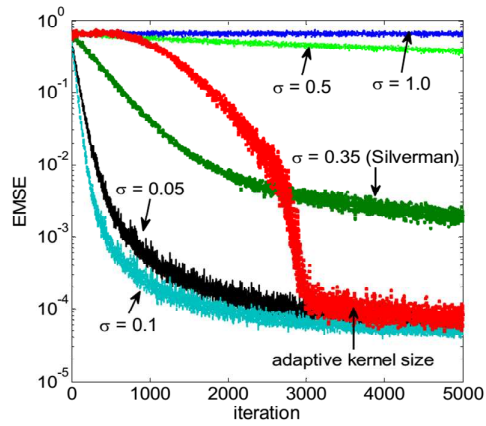}
\caption{Convergence curves for different kernel sizes}
\label{fig:example}
\end{figure}

\begin{figure}
\centering
\includegraphics[height=5.5cm]{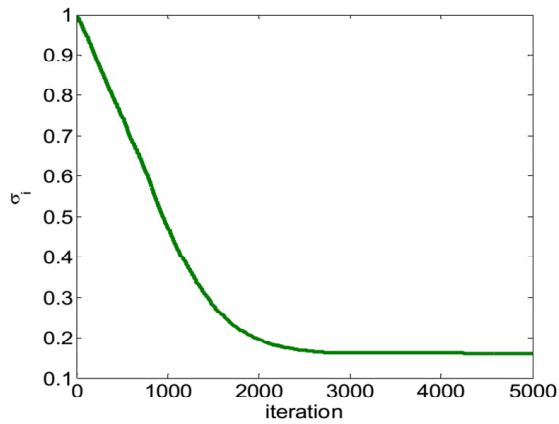}
\caption{Evolution of the adaptive kernel size $\sigma_i$}
\label{fig:example}
\end{figure}

\begin{figure}
\centering
\includegraphics[height=5.5cm]{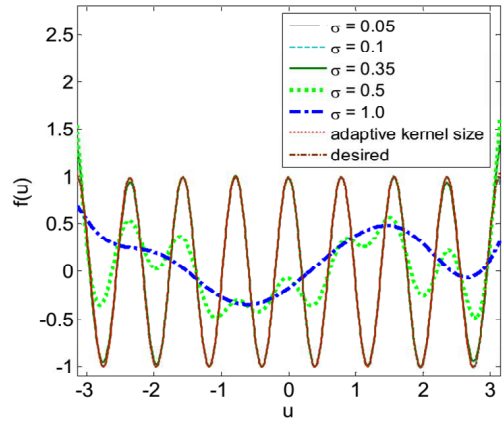}
\caption{Learned mappings at final iteration}
\label{fig:example}
\end{figure}

\begin{figure}
\centering
\includegraphics[height=5.5cm]{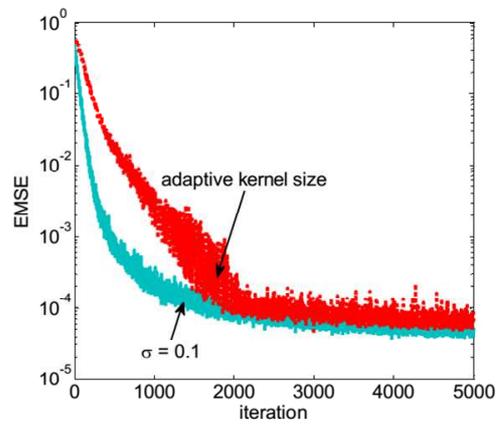}
\caption{Convergence curves for $\sigma=0.1$ and the adaptive kernel size with initial value determined by Silverman¡¯s rule}
\label{fig:example}
\end{figure}

\subsection{Short Term Chaotic Time Series Prediction}
The second example is about short term chaotic time series prediction. Consider the Lorenz oscillator whose state equations are
\begin{equation}
\left\{ \begin{array}{l}
\frac{\displaystyle{dx}}{\displaystyle{dt}} =  - \beta x + yz\\
\frac{\displaystyle{dy}}{\displaystyle{dt}} = \delta (z - y)\\
\frac{\displaystyle{dz}}{\displaystyle{dt}} =  - xy + \rho y - z
\end{array} \right.
\end{equation}
where the parameters are set as $\beta  = 4.0$, $\delta=30$, and $\rho=45.92$. The second state is picked for the prediction task and a sample time series is shown in Fig. 5. Here the goal is to predict the value of the current sample $y(i)$ using the previous five consecutive samples.

We continue to compare the performances of the KLMS with different kernel sizes. In the simulations below, the step-sizes for the mapping update are set at $\eta=0.1$, and the step-size for the kernel size adaptation is set at $\rho=0.05$. The initial value of the adaptive kernel size $\sigma_i$ is set as 1.0. For the kernel sizes $\sigma=1.0$, 5.5(selected by Silverman¡¯s rule), 10, 15, 20, 30 and the adaptive kernel size $\sigma_i$, the convergence curves in terms of the testing MSE are illustrated in Fig. 6. For each kernel size, 20 independent simulations were run with different segments of the time series. In each segment, 1000 samples are used as the training data and another 100 as the test data. At each iteration, the testing MSE is computed based on the test data using the learned filter at that iteration. It can be seen from Fig. 6 that the kernel size $\sigma=15$ achieves the best performance with the smallest testing MSE at final iteration. Also, as expected, the adaptive kernel size yields a satisfactory performance very close to the best result.  Fig. 7 shows the evolution of the adaptive kernel size (see the solid line). Interestingly, we observe the adaptive kernel size $\sigma_i$ converges quickly and very close to the desirable value 15. The mean deviation results of the testing MSE at final iteration are given in Table 3.

\begin{figure}
\centering
\includegraphics[height=5.5cm]{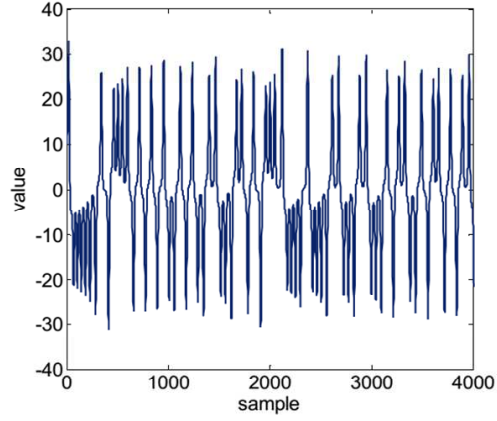}
\caption{Time series produced by Lorenz system}
\label{fig:example}
\end{figure}

\begin{figure}
\centering
\includegraphics[height=5.5cm]{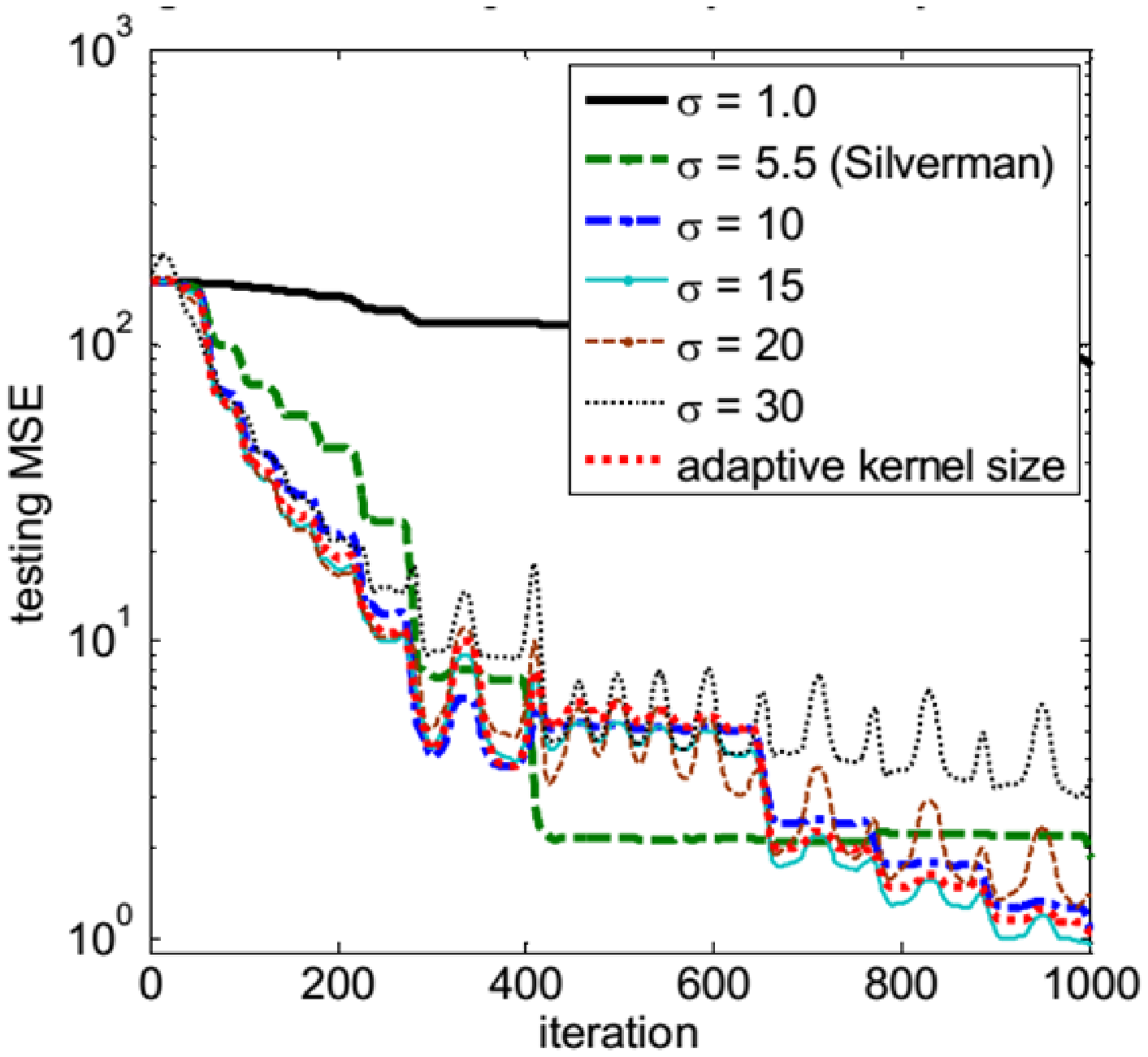}
\caption{Convergence curves in terms of the testing MSE for different kernel sizes}
\label{fig:example}
\end{figure}

\begin{figure}
\centering
\includegraphics[height=5.5cm]{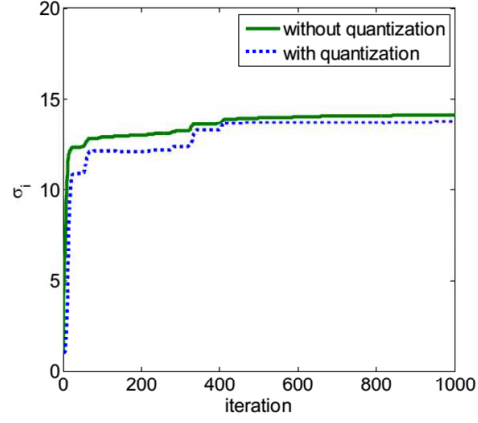}
\caption{Evolution of the kernel size $\sigma_i$ in Lorenz time series prediction}
\label{fig:example}
\end{figure}

\begin{figure}
\centering
\includegraphics[height=5.5cm]{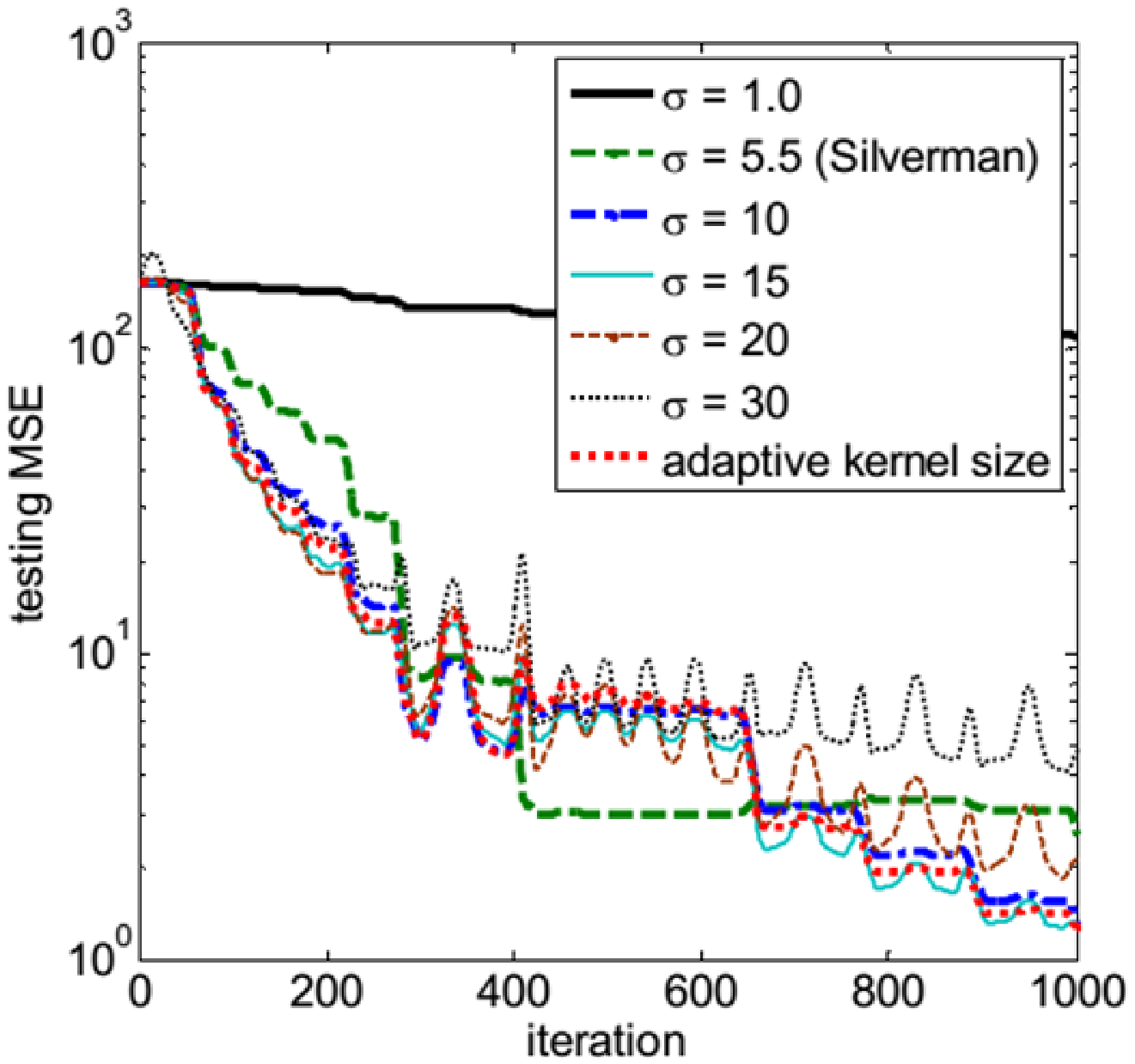}
\caption{Convergence curves in terms of the testing MSE for different kernel sizes (with quantization)}
\label{fig:example}
\end{figure}

\begin{figure}
\centering
\includegraphics[height=5.5cm]{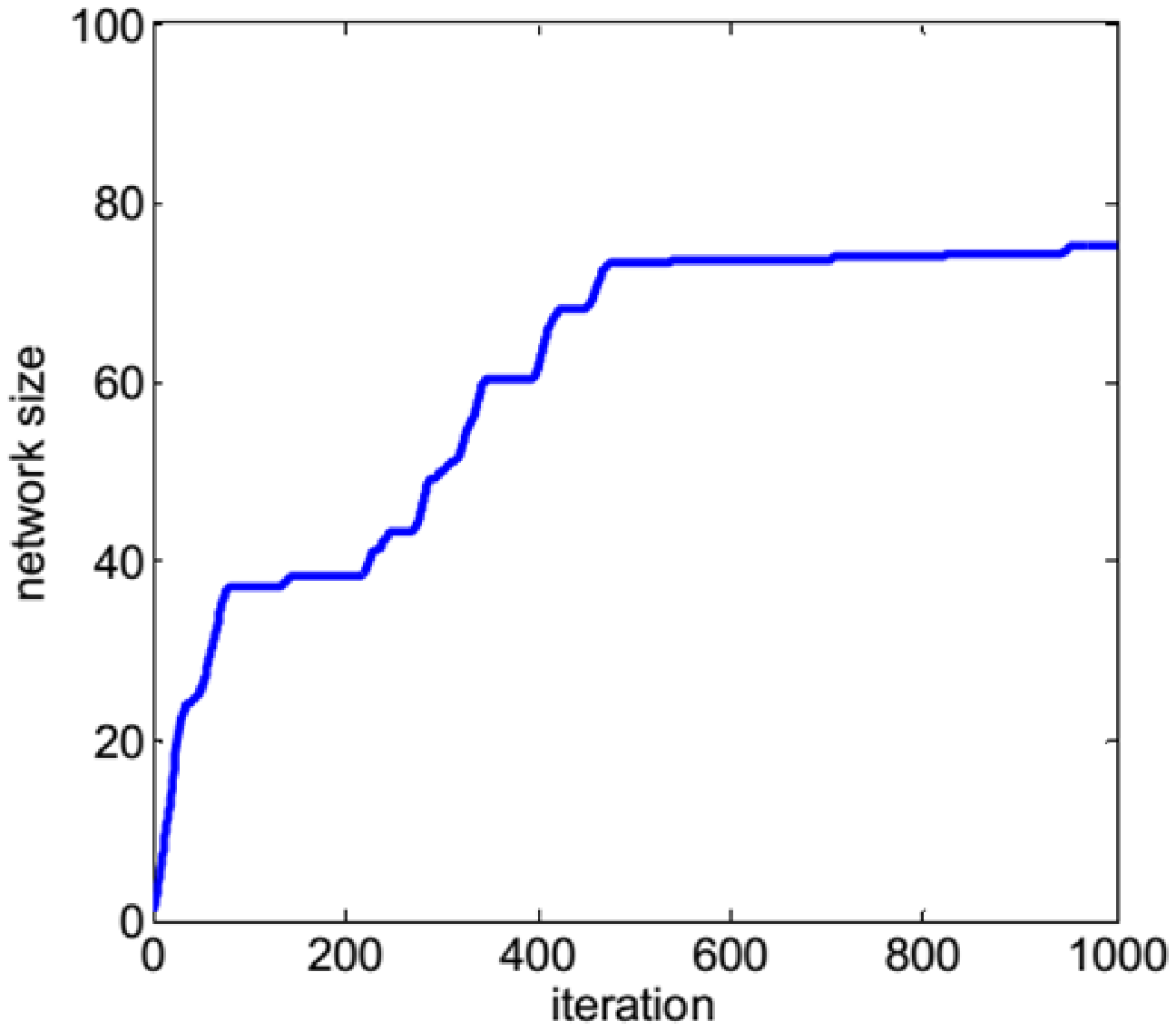}
\caption{Network size evolution in Lorenz time series prediction (with quantization)}
\label{fig:example}
\end{figure}

Further, we would like to evaluate the performance when the quantization method is applied to curb the growth of the network size. The experimental setting is the same as before, except now the quantization method is used and the quantization size is set at $\varepsilon  = 4.0$. The convergence curves for different kernel sizes are demonstrated in Fig. 8. Once again, the adaptive kernel size works very well, and obtains a performance very close to the best one. As shown in Fig. 7, the adaptive kernel size $\sigma_i$ still converges close to the value 15 (see the dotted line). Fig. 9 illustrates the evolution curve of the network size. One can see that with quantization the network size grows very slowly, and the final network size is only around 75. Table 4 shows the mean deviation results of the testing MSE at final iteration.

\begin{table}
\begin{center}
\begin{tabular}{c|c}
\hline

\hline\noalign{\smallskip}
Kernel Size & Testing MSE\\
\hline
$\sigma=1.0 $ & $85.3740 \pm 2.6436$ \\
\hline
$\sigma=5.5 $ & $1.8477 \pm 0.2751$ \\
\hline
$\sigma=10 $ & $1.0450 \pm 0.1338$ \\
\hline
$\sigma=15 $ & $0.9490 \pm 0.0276$ \\
\hline
$\sigma=20 $ & $1.3876 \pm 0.0281$ \\
\hline
$\sigma=30 $ & $3.4480 \pm 0.1370$ \\
\hline
adaptive kernel size & $1.0264 \pm 0.1018$ \\
\hline

\hline
\noalign{\smallskip}
\end{tabular}
\caption{Testing MSE at final iteration}
\end{center}
\end{table}

\begin{table}
\begin{center}
\begin{tabular}{c|c}
\hline

\hline\noalign{\smallskip}
Kernel Size & Testing MSE\\
\hline
$\sigma=1.0 $ & $109.2801 \pm 1.1974$ \\
\hline
$\sigma=5.5 $ & $2.5486 \pm 0.4112$ \\
\hline
$\sigma=10 $ & $1.3183 \pm 0.1509$ \\
\hline
$\sigma=15 $ & $1.2982 \pm 0.0262$ \\
\hline
$\sigma=20 $ & $2.1173 \pm 0.0215$ \\
\hline
$\sigma=30 $ & $4.8447 \pm 0.2272$ \\
\hline
adaptive kernel size & $ 1.2626 \pm 0.0890$ \\
\hline

\hline
\noalign{\smallskip}
\end{tabular}
\caption{Testing MSE at final iteration (with quantization)}
\end{center}
\end{table}

\section{Conclusion}
The kernel function implicitly defines the feature space and plays a central role in all kernel methods. In kernel adaptive filtering (KAF) algorithms, the Gaussian kernel (a radial basis function kernel) is usually a default kernel. The kernel size (or bandwidth) of the Gaussian kernel controls the smoothness of the mapping and has significant influence on the learning performance. How to select a proper kernel size is a very crucial problem in KAF algorithms. Some existing techniques (e.g. Silverman¡¯s rule) for selecting a kernel size can be applied, but they are not appropriate for a KAF algorithm since the problem is approximation in a joint space (the input and the desired), which is different from density estimation.

In this work, we propose an approach for sequentially optimizing the kernel size for the kernel least mean square (KLMS), a simple yet efficient KAF algorithm. At each iteration cycle, the kernel size is adjusted by a stochastic gradient based algorithm to minimizing the mean square error. The proposed algorithm is computationally very simple and easy to implement. Theoretical results on convergence are also presented. Based on the energy conservation relation in RKHS, we derive a sufficient condition for the mean square convergence, and obtain the theoretical steady-state excess mean-square error (EMSE). Simulation results confirm the theoretical prediction, and show the adaptive kernel size can automatically converge to a proper value, so as to help KLMS converge faster and achieve better accuracy.

In future study, it is of interest to extend this work to the case where the kernel is of any form (not restricted to the Gaussian kernel). Especially, we will study how to sequentially optimize the kernel function using the idea of multi-kernel learning or learning the kernel. Another interesting line of study is how to jointly optimize the kernel size and the step size.

\bibliographystyle{splncs}
\bibliography{typeinst_cbd}

\end{document}